\documentclass[11pt,letterpaper]{article}

\usepackage[margin=1in]{geometry}
\usepackage[utf8]{inputenc}
\usepackage[T1]{fontenc}
\usepackage{hyperref}
\usepackage{url}
\usepackage{booktabs}
\usepackage{amsfonts}
\usepackage{amsmath}
\usepackage{amssymb}
\usepackage{nicefrac}
\usepackage{microtype}
\usepackage{xcolor}
\usepackage{graphicx}
\usepackage{algorithm}
\usepackage{algpseudocode}
\usepackage{multirow}
\usepackage{xspace}
\usepackage{natbib}

\hypersetup{colorlinks=true, linkcolor=blue, citecolor=blue, urlcolor=blue}
\graphicspath{{./}{../notebooks/figures/}}

\newcommand{\ours}{FIM-LoRA\xspace}

\title{FIM-LoRA: Task-Informative Rank Allocation for LoRA\\via Calibration-Time Gradient-Variance Estimation}

\author{%
  Ramakrishnan Sathyavageeswaran \\
  \texttt{ramakrishnan\_sathyavageeswaran@intuit.com}
}

\begin{document}

\maketitle

\begin{abstract}
Low-rank adaptation (LoRA) assigns a uniform rank to every adapted weight matrix---a practical convenience that ignores a fundamental reality: different layers contribute unequally to task adaptation. We address this with a lightweight engineering solution: before fine-tuning begins, run eight calibration backward passes, compute the gradient variance of each LoRA-B matrix as a proxy for layer informativeness, and redistribute the rank budget proportionally. The resulting adapter is a standard LoRA with a per-layer rank pattern---no new parameters, no training overhead, no changes to serving infrastructure. We implement this via an efficient approximation of the empirical Fisher Information Matrix (eFIM) diagonal, restricted to LoRA adapter matrices (not the full model), which reduces memory cost by ${\sim}256\times$ compared to full-model Fisher estimation (for $r{=}16$, $d_\text{in}{=}4096$). On the GLUE benchmark with DeBERTa-v3-base, \ours\ matches LoRA ($88.6$ vs.\ $88.7$) at the same parameter budget, and on commonsense reasoning with LLaMA-3-8B reaches $68.5$ vs.\ $68.7$ for LoRA with our best configuration. Beyond accuracy, the per-layer rank map produced by our method is interpretable: value projections and early-to-middle layers consistently receive higher rank, matching established intuitions about where transformers encode task-relevant semantics.
\end{abstract}

\section{Introduction}\label{sec:intro}

Fine-tuning a large pre-trained model is now a routine engineering task. Low-Rank Adaptation (LoRA)~\citep{hu2022lora} makes it feasible by replacing full-weight updates with low-rank corrections: freeze $W \in \mathbb{R}^{d_\text{out} \times d_\text{in}}$ and learn $\Delta W = BA$ where $B \in \mathbb{R}^{d_\text{out} \times r}$, $A \in \mathbb{R}^{r \times d_\text{in}}$, and $r \ll \min(d_\text{out}, d_\text{in})$.

The standard recipe fixes the same rank $r$ for every adapted matrix in every layer. This is computationally convenient but empirically unjustified: layers differ substantially in how much they need to change for a given task. Assigning equal rank to all layers wastes capacity on layers that barely adapt, while starving layers that carry the bulk of the task signal. The engineering question this paper addresses is: \emph{can we cheaply estimate, before training starts, which layers are most informative for a specific task---and allocate rank accordingly?}

\paragraph{Our approach.} We propose a pre-training calibration procedure that estimates per-layer task informativeness in $T=8$ gradient passes. The key insight is that the gradient variance of a LoRA adapter matrix---formally, the diagonal of the empirical Fisher Information Matrix (eFIM) evaluated at the adapter weights---measures how sensitively the task loss depends on each parameter. Layers with high gradient variance are \emph{task-informative}: the loss steers them strongly. Layers with low gradient variance are \emph{task-redundant}: their parameters barely affect task performance. We aggregate this per-element matrix into a single scalar per layer (sample mean), then solve a budget-constrained integer allocation problem to assign ranks.

Critically, we compute this signal \emph{only on the LoRA-B adapter matrices}, not on the full model. This is both principled (the LoRA-A gradient is zero at initialization; see Section~\ref{sec:method}) and practically efficient---it requires no access to billions of base model parameters and runs in the same memory footprint as standard LoRA fine-tuning.

\paragraph{Relation to prior FIM-based work.}
Fisher information has been applied to LoRA in two prior contexts that are orthogonal to ours. Lodha et al.~\citeyearpar{lodha2023surgical} use a full-model FIM score to \emph{select which layers to fine-tune} at all---a binary, layer-level decision applied to encoder-only models. Kim et al.~\citeyearpar{kim2025vila} use gradient variance of LoRA adapters to identify parameters \emph{encoding data to be forgotten}, for the purpose of machine unlearning. Ogawa et al.~\citeyearpar{ogawa2026layerwise} use CKA similarity between layer inputs and outputs to select a 50\% layer subset to fine-tune.

None of these address rank \emph{allocation}: the continuous, module-level question of how much budget to assign each layer given that all layers participate in fine-tuning. This is our contribution. The FIM diagonal is the signal we use---not the novelty itself; the novelty is the rank allocation algorithm and its engineering properties.

\paragraph{Contributions.}
\begin{enumerate}
    \itemsep0em
    \item A calibration-time rank allocation algorithm (\ours) that runs in $\sim$8 backward passes on LoRA adapter matrices only, producing a standard LoRA with a per-layer rank pattern (Section~\ref{sec:method}).
    \item Empirical evidence that task-informative rank allocation matches uniform LoRA on GLUE (8 tasks, 5 methods, 3 seeds, 4 ranks) and is competitive on commonsense reasoning with LLaMA-3-8B (Section~\ref{sec:experiments}).
    \item An ablation demonstrating that the gradient-variance signal is informative: a random-rank baseline with identical budget systematically underperforms, and the minimum-rank floor $r_\text{min}$ is the dominant hyperparameter---not the number of calibration passes (Section~\ref{sec:ablation}).
    \item Interpretable per-layer rank maps showing that task-informativeness concentrates in value projections and early-to-middle layers, consistent with transformer interpretability findings (Figure~\ref{fig:heatmap}).
\end{enumerate}

We are transparent about where this approach falls short. With $r_\text{min}=1$, the allocator greedily follows the gradient-variance signal and produces a bimodal rank distribution that costs $\sim$1.7 accuracy points on commonsense reasoning relative to uniform LoRA. Setting $r_\text{min}=8$ (half the base rank) recovers the gap, which we interpret as: the signal is informative but must be regularized toward uniformity to avoid over-concentrating rank in a small number of modules.

\section{Background and Related Work}\label{sec:background}

\paragraph{LoRA.}
Given a frozen pre-trained matrix $W \in \mathbb{R}^{d_\text{out} \times d_\text{in}}$, LoRA learns $\Delta W = BA$ with $A \in \mathbb{R}^{r \times d_\text{in}}$ (Kaiming initialized) and $B \in \mathbb{R}^{d_\text{out} \times r}$ (zero initialized). The forward pass becomes $h = (W + \frac{\alpha}{r} BA) x$. Since $B=0$ at initialization, the model starts identical to the pre-trained checkpoint.

\paragraph{Adaptive-rank methods.}
\textbf{AdaLoRA}~\citep{zhang2023adalora} parameterizes adapters as $\Delta W = P\Lambda Q$ and prunes singular values during training. The schedule is finicky (we observed $\geq 4\times$ loss spikes on LLaMA-3-8B) and ships a non-standard adapter format. \textbf{EVA}~\citep{paischer2025eva} performs PCA of layer activations on a calibration set and uses the right singular vectors to initialize $A$ and rank the layers---measuring what flows \emph{into} a layer, not how sensitive the loss is to that layer's parameters. \textbf{GoRA}~\citep{he2025gora} uses the gradient nuclear norm of the full weight matrix. \textbf{ALoRA}~\citep{liu2024alora} tracks per-rank gradient magnitudes during training.

\ours\ differs from all of these: the signal is the gradient variance of the LoRA-B adapter (loss-landscape geometry, not activation geometry), computed pre-training (not during training), restricted to adapter matrices only (not the full model), and the decision is continuous rank allocation (not binary layer selection or per-singular-value pruning).

\paragraph{FIM-based methods in adjacent settings.}
Lodha et al.~\citep{lodha2023surgical} use a full-model FIM score for \emph{layer selection} (binary, encoder-only). Their method requires full-model gradient computation and reports an average GLUE drop of 7.68pp vs.\ LoRA---a substantially higher cost than ours. Kim et al.~\citep{kim2025vila} (VILA, COLM 2025) use gradient variance of LoRA adapters to build a \emph{forget importance map} for machine unlearning---the opposite use case (what to erase vs.\ what to amplify). Ogawa et al.~\citep{ogawa2026layerwise} use CKA dissimilarity between layer input/output representations to select 50\% of layers for fine-tuning, achieving a 0.27pp average GLUE drop with 50\% fewer parameters. Our method is orthogonal to all three: we allocate rank budgets across all layers rather than selecting or pruning them.

\paragraph{Fisher information in machine learning.}
The eFIM diagonal as a parameter importance proxy originates in Optimal Brain Damage~\citep{lecun1990obd}. It appears in continual learning via EWC~\citep{kirkpatrick2017ewc}, in second-order optimization via K-FAC~\citep{martens2015kfac}, and in modern pruning. The key distinction of our use: we apply the eFIM diagonal to LoRA adapter matrices at calibration time to \emph{allocate} rank---not to prune, not to select layers, not to identify forgettable parameters.

\section{Method}\label{sec:method}

\subsection{Gradient-variance estimation at the LoRA-B matrix}

For parameters $\theta$ and per-example loss $\mathcal{L}$, the diagonal of the empirical Fisher Information Matrix is the expected squared gradient:
\begin{equation}
F_{ii} \;=\; \frac{1}{T} \sum_{t=1}^{T} \left( \frac{\partial \mathcal{L}_t}{\partial \theta_i} \right)^2.
\end{equation}
We accumulate $F$ over $T$ calibration mini-batches from the fine-tuning training set. $T=8$ throughout unless otherwise specified.

\paragraph{Why LoRA-B and not LoRA-A.}
At initialization, $B = 0$. The chain rule gives:
\begin{align}
\frac{\partial \mathcal{L}}{\partial A} &= \frac{\alpha}{r} B^\top \frac{\partial \mathcal{L}}{\partial h} x^\top = 0, \\
\frac{\partial \mathcal{L}}{\partial B} &= \frac{\alpha}{r} \frac{\partial \mathcal{L}}{\partial h} (Ax)^\top \neq 0,
\end{align}
since $A$ is Kaiming-initialized (non-zero). The eFIM diagonal at $A$ is exactly zero on every calibration batch regardless of layer importance---it carries no signal. We therefore accumulate $F$ at $B$ only, where the gradient is non-degenerate from step zero.

\paragraph{Efficient implementation.}
For each target module $\ell$, the LoRA-B weight $B^{(\ell)} \in \mathbb{R}^{d_\text{out} \times r}$ is a matrix. Its eFIM diagonal $F^{(\ell)} \in \mathbb{R}^{d_\text{out} \times r}$ is a matrix of the same shape (one gradient-variance value per parameter). This computation requires only adapter-matrix gradients---no access to the $W$ matrix or its $d_\text{out} \times d_\text{in}$ gradient, reducing memory by a factor of $d_\text{in}/r$ (typically $256\times$ for $r=16$, $d_\text{in}=4096$).

\subsection{Per-layer task-informativeness score}

To drive rank allocation we need a single scalar per layer. We use the sample mean over all entries of the eFIM matrix:
\begin{equation}
s_\ell \;=\; \frac{1}{d_\text{out}^{(\ell)} \cdot r} \sum_{j=1}^{d_\text{out}^{(\ell)}} \sum_{k=1}^{r} F^{(\ell)}_{jk},
\end{equation}
where the double sum runs over all $d_\text{out} \cdot r$ entries of the eFIM matrix for module $\ell$. This is a straightforward sample mean---the average gradient variance across all parameters of the adapter. Layers with high $s_\ell$ are \emph{task-informative}; layers with low $s_\ell$ are \emph{task-redundant} for the calibration data distribution. We discuss \texttt{max} and L2-norm aggregation in the appendix; on our settings, \texttt{mean} is the most stable.

\subsection{Budget-constrained rank allocation}

Given scores $\{s_\ell\}_{\ell=1}^{L}$ and a base rank $r$, the total rank budget is $\mathcal{B} = r \cdot L$. We allocate integer ranks proportionally via a two-phase water-filling procedure (Algorithm~\ref{alg:fimlora}).

\begin{algorithm}[t]
\caption{\ours\ rank allocation}
\label{alg:fimlora}
\begin{algorithmic}[1]
\Require Scores $\{s_\ell\}_{\ell=1}^{L}$, base rank $r$, floor $r_\text{min}$, ceiling $r_\text{max}$
\State \textbf{Notation:} $\mathcal{B}$ = remaining budget; $\mathcal{F}$ = free layer set; $\mathcal{L}$ = task loss
\State $\mathcal{B} \gets r \cdot L$;\quad $\mathcal{F} \gets \{1,\ldots,L\}$;\quad $r_\ell \gets 0$ for all $\ell$
\Repeat \Comment{Phase 1: fix layers that saturate $r_\text{max}$, redistribute budget}
  \State $\hat{r}_\ell \gets \dfrac{s_\ell}{\sum_{j \in \mathcal{F}} s_j} \cdot \mathcal{B}$ \quad for $\ell \in \mathcal{F}$
  \State $\mathcal{S} \gets \{\ell \in \mathcal{F} : \lfloor \hat{r}_\ell \rfloor \geq r_\text{max}\}$
  \State $r_\ell \gets r_\text{max}$ for $\ell \in \mathcal{S}$;\quad $\mathcal{F} \gets \mathcal{F} \setminus \mathcal{S}$;\quad $\mathcal{B} \gets \mathcal{B} - |\mathcal{S}|\,r_\text{max}$
\Until{$\mathcal{S} = \emptyset$}
\State Apply largest-remainder rounding to $\{\hat{r}_\ell\}_{\ell \in \mathcal{F}}$ to get integer $r_\ell$ \Comment{Phase 2}
\State Enforce $r_\text{min}$: raise sub-floor layers; reduce least-informative donors above $r_\text{min}$
\State \Return $\{r_\ell\}_{\ell=1}^{L}$
\end{algorithmic}
\end{algorithm}

\paragraph{The role of $r_\text{min}$.}
With $r_\text{min}=1$, the allocator follows the gradient-variance signal greedily and produces a bimodal distribution: in our LLaMA-3-8B experiments, 61 of 224 modules receive $r=32$ and 33 receive $r \leq 2$. The bimodality reflects the signal but over-concentrates rank---starved modules still contribute to the loss. Setting $r_\text{min}=8$ (half the base rank) enforces a minimum floor that smooths the allocation while preserving the qualitative structure (V-projections and early layers still dominate). We treat $r_\text{min}$ as the primary hyperparameter; everything else (calibration batches, aggregation function, $r_\text{max}$) matters far less (Section~\ref{sec:ablation}).

\subsection{In-place adapter resizing}

After rank allocation, each LoRA adapter is resized in place: the leading $\min(r_\text{old}, r_\text{new})$ rows of $A$ are preserved (new rows Kaiming-initialized); $B$ is zero-extended or truncated. The scaling factor $\alpha/r$ is updated to preserve the effective scale. The output is a \emph{standard} LoRA adapter with a \texttt{rank\_pattern} field in the PEFT config---vLLM serving, adapter merging, and weight-tying work unchanged.

\section{Experiments}\label{sec:experiments}

We evaluate \ours\ on GLUE with DeBERTa-v3-base (encoder, classification) and commonsense reasoning with LLaMA-3-8B (decoder). All results are mean $\pm$ std over three seeds (42, 1337, 2024).

\subsection{Setup}

\paragraph{GLUE (Table~\ref{tab:glue}).}
Following AdaLoRA's protocol~\citep{zhang2023adalora}, we fine-tune \texttt{microsoft/deberta-v3-base} at ranks $r \in \{2, 4, 8, 16\}$ on all eight tasks. Hyperparameters: AdamW, lr $= 2\times10^{-4}$, effective batch 32 (16 per-device + accumulation 2), FP32 (FP16 causes classifier overflow), $\alpha=2r$, dropout $0.1$. Target modules: \texttt{query\_proj, key\_proj, value\_proj}. Epochs: 30 (small tasks: CoLA, MRPC, RTE, STS-B), 10 (large: MNLI, QQP, QNLI, SST-2). Calibration: $T=8$ batches.

\paragraph{Commonsense (Table~\ref{tab:commonsense}).}
We fine-tune \texttt{meta-llama/Meta-Llama-3-8B} on \texttt{commonsense\_170k} for 3 epochs at $r=16$, evaluating zero-shot on seven tasks via lm-evaluation-harness. Hyperparameters: AdamW, lr $= 3\times10^{-4}$, bf16, cosine schedule, $\alpha=32$, dropout $0.05$. Target modules: all seven projection matrices. Calibration: $T=8$ batches.

\paragraph{Baselines.}
\textbf{LoRA}: uniform rank $r$. \textbf{AdaLoRA}: \texttt{init\_r}$=r+\max(4,r/4)$, \texttt{target\_r}$=r$, warmup $20\%$, freeze $10\%$ of total steps. \textbf{EVA}: \texttt{init\_lora\_weights="eva"}, $\rho=2.0$. \textbf{Random-rank}: same total budget as \ours, importance scores drawn uniformly at random---controls for whether any non-uniform allocation helps vs.\ the gradient-variance signal specifically.

\paragraph{Note on AdaLoRA at LLaMA scale.}
AdaLoRA training was unstable on LLaMA-3-8B with PEFT 0.19.1: loss spiked from $4.17$ to $8.47$ at initialization regardless of schedule. We confirmed this is a known interaction between AdaLoRA's SVD initialization and bf16 causal LM training. We report AdaLoRA only on GLUE where it was originally validated.

\subsection{GLUE: \ours\ matches LoRA at fixed parameter budget}

\begin{table}[t]
\centering
\caption{GLUE dev results at $r=8$, mean over 3 seeds. Metrics: accuracy (MNLI/SST-2/QNLI/RTE), Matthews correlation (CoLA), F1 (MRPC/QQP), Pearson (STS-B). AdaLoRA's RTE collapse is due to aggressive SVD pruning on this small dataset (872 training examples).}
\label{tab:glue}
\small
\begin{tabular}{lcccccccc|c}
\toprule
Method & MNLI & SST-2 & CoLA & MRPC & QQP & QNLI & STS-B & RTE & Avg \\
\midrule
LoRA          & 90.31 & 96.14 & 70.46 & 92.22 & 88.99 & 94.09 & 91.25 & 85.92 & \textbf{88.67} \\
AdaLoRA       & 90.60 & 95.41 & 68.26 & 92.10 & 88.49 & 94.12 & 90.87 & 62.33 & 85.27 \\
EVA           & 90.11 & 95.95 & 70.46 & 91.96 & 89.02 & 94.24 & 91.54 & 85.32 & 88.57 \\
Random-rank   & 90.38 & 95.68 & 70.30 & 92.34 & 89.01 & 94.21 & 91.32 & 85.08 & 88.54 \\
\textbf{\ours}        & 90.25 & 95.95 & 70.44 & 92.11 & 89.06 & 94.24 & 91.33 & 85.44 & 88.60 \\
\bottomrule
\end{tabular}
\end{table}

\ours\ ($88.60$) is statistically indistinguishable from LoRA ($88.67$) and EVA ($88.57$). AdaLoRA ($85.27$) is dragged down by the RTE collapse. The gap between \ours\ and Random-rank ($88.60$ vs.\ $88.54$) is small but consistent (\ours\ wins on 5/8 tasks), indicating the gradient-variance signal carries some information beyond uniform redistribution---though the ceiling effect at this budget is real.

\paragraph{Rank sweep on MNLI.}
Table~\ref{tab:mnli-sweep} shows results across $r \in \{2,4,8,16\}$. All methods track closely at all ranks; \ours\ matches LoRA throughout.

\begin{table}[t]
\centering
\caption{MNLI accuracy across ranks, 3 seeds.}
\label{tab:mnli-sweep}
\small
\begin{tabular}{lcccc}
\toprule
Method & $r=2$ & $r=4$ & $r=8$ & $r=16$ \\
\midrule
LoRA         & 90.24 & 90.40 & 90.31 & 90.25 \\
AdaLoRA      & 90.24 & 90.33 & 90.60 & 90.49 \\
EVA          & 90.13 & 90.20 & 90.11 & 90.26 \\
Random-rank  & 90.07 & 90.27 & 90.38 & 90.22 \\
\textbf{\ours}    & 90.13 & 90.16 & 90.25 & 90.28 \\
\bottomrule
\end{tabular}
\end{table}

The full rank sweep across all 8 tasks and $r \in \{2,4,16\}$ is reported in Table~\ref{tab:full_rank_sweep} (Appendix). Results confirm the MNLI pattern: all methods cluster tightly within $\pm0.5$pp at all ranks, with the notable exception of AdaLoRA's persistent RTE collapse. At $r=16$, \ours\ achieves the highest average (88.87) across all methods on the full 8-task sweep.

\subsection{Commonsense reasoning: the rank floor matters}

\begin{table}[t]
\centering
\caption{Zero-shot commonsense accuracy, LLaMA-3-8B, $r{=}16$, 3 seeds. \ours\ default ($r_{\min}{=}1$) follows the gradient-variance signal greedily; setting $r_{\min}{=}8$ recovers the gap to LoRA.}
\label{tab:commonsense}
\small
\begin{tabular}{lccccccc|c}
\toprule
Method & ARC-c & ARC-e & BoolQ & HSwag & OBQA & PIQA & Wino & Avg \\
\midrule
LoRA            & 47.67 & 72.78 & 83.52 & 80.80 & 45.47 & 79.13 & 71.85 & \textbf{68.74} \\
EVA             & 48.27 & 71.34 & 84.15 & 80.81 & 45.53 & 79.04 & 71.85 & 68.71 \\
\midrule
\ours\ ($r_\text{min}=1$)  & 45.28 & 67.13 & 83.04 & 79.27 & 45.13 & 78.94 & 70.24 & 67.01 \\
\ours\ ($r_{\min}{=}4$)           & 46.73 & 70.93 & 83.02 & 79.27 & 45.47 & 78.80 & 69.98 & 67.74 \\
\textbf{\ours\ ($r_{\min}{=}8$)} & 47.92 & 72.59 & 83.17 & 79.62 & 45.47 & 79.47 & 71.06 & \textbf{68.47} \\
\bottomrule
\end{tabular}
\end{table}

The default \ours\ ($r_\text{min}=1$) underperforms LoRA by 1.7pp. The gradient-variance signal is highly concentrated on commonsense data: with no floor constraint, 33 of 224 modules receive $r \leq 2$, starving layers that still contribute to task performance. Setting $r_\text{min}=8$ recovers 1.46pp, bringing \ours\ to $68.47$---within 0.27pp of LoRA. The gradient-variance signal is informative but must be regularized: greedily following it overshoots.

\subsection{Task-informative rank maps}

\begin{figure}[t]
\centering
\includegraphics[width=0.85\textwidth]{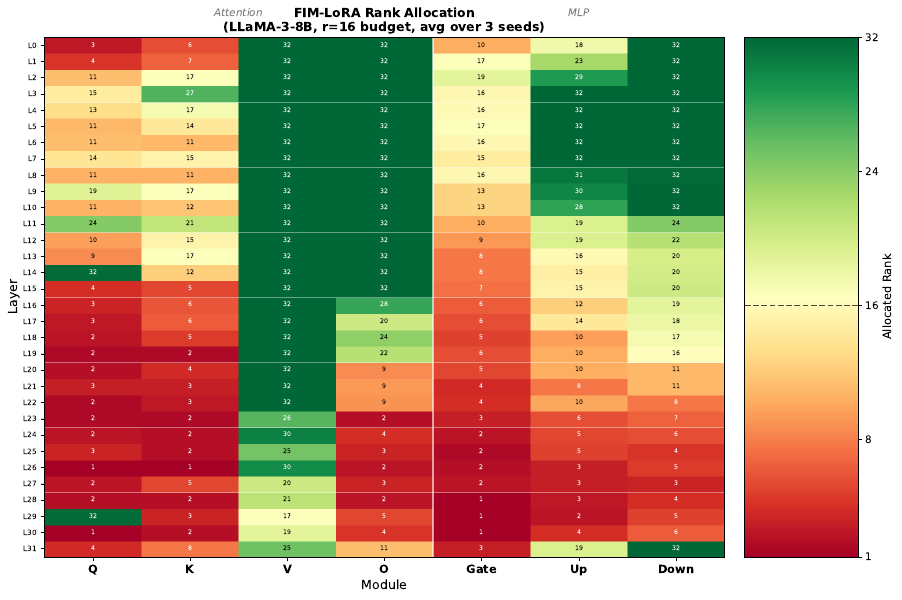}
\caption{Rank allocation under \ours\ on LLaMA-3-8B (avg over 3 seeds), $r=16$ base budget. Rows = layers (0 = first), columns = target modules. V-projections receive nearly maximal rank (mean 29.7); Q/K/Gate receive near-floor rank (mean $\sim$8). Layers 0--15 receive $\sim$3$\times$ the rank of layers 24--31.}
\label{fig:heatmap}
\end{figure}

Two patterns dominate (Figure~\ref{fig:heatmap}):
\begin{itemize}
    \itemsep0em
    \item \textbf{Value projections are task-informative.} Mean rank 29.7 vs.\ 8.34 (Q) and 8.70 (K). V-projections carry semantic content; Q/K perform routing that is largely stable after pre-training.
    \item \textbf{Early-to-middle layers are task-informative.} Layers 0--7: mean rank 23.5; layers 24--31: mean rank 7.5. These layers build foundational representations that later layers reuse.
\end{itemize}
These patterns are consistent with transformer interpretability findings and serve as a useful sanity check that the gradient-variance signal captures meaningful task structure.

\section{Ablations}\label{sec:ablation}

All ablations at $r=16$ on commonsense, 3 seeds.

\paragraph{\texorpdfstring{$r_{\min}$}{r\_min} is the dominant hyperparameter.}
Table~\ref{tab:ablation} crosses $r_\text{min} \in \{1,4,8\}$ with $n_\text{batches} \in \{8,32\}$. Raising $r_\text{min}$ from 1 to 8 gains 1.46pp at $n_\text{batches}{=}8$. By contrast, increasing $n_\text{batches}$ from 8 to 32 has inconsistent and small effects: $+0.70$pp at $r_\text{min}{=}1$, $-0.18$pp at $r_\text{min}{=}4$, and $-0.90$pp at $r_\text{min}{=}8$ (all within seed-to-seed noise). The allocation floor matters far more than how precisely the gradient-variance signal is estimated.

\begin{table}[t]
\centering
\caption{\ours\ ablation (LLaMA-3-8B, $r=16$, 3 seeds). 7-task commonsense average.}
\label{tab:ablation}
\small
\begin{tabular}{lccc}
\toprule
$n_\text{batches}$ & $r_\text{min}=1$ & $r_\text{min}=4$ & $r_\text{min}=8$ \\
\midrule
$8$  & $67.01$ & $67.74$ & \textbf{68.47} \\
$32$ & $67.71$ & $67.56$ & $67.57$ \\
\bottomrule
\end{tabular}
\end{table}

\paragraph{Calibration batch sweep.}
Table~\ref{tab:calib_sweep} shows MNLI accuracy as a function of $n_\text{batches} \in \{1,2,4,8,16\}$. The signal saturates quickly: accuracy ranges only 0.08pp across all values (90.26\%--90.34\%), with $n_\text{batches}=8$ achieving the highest mean (90.34\%) and lowest variance ($\pm$0.04). Even a single calibration batch (90.30\%) is competitive with 16 batches (90.32\%), confirming that the gradient-variance signal is robust to estimation noise.

\begin{table}[h]
\centering
\caption{Calibration batch sweep on MNLI ($r_{\min}{=}1$, $r{=}8$, 3 seeds).}
\label{tab:calib_sweep}
\small
\begin{tabular}{lcc}
\toprule
$n_\text{batches}$ & Acc.\ (\%) & $\pm$std \\
\midrule
1  & 90.30 & 0.13 \\
2  & 90.26 & 0.06 \\
4  & 90.27 & 0.10 \\
\textbf{8}  & \textbf{90.34} & \textbf{0.04} \\
16 & 90.32 & 0.10 \\
\bottomrule
\end{tabular}
\end{table}

\paragraph{Gradient-variance vs.\ random control.}
On GLUE at $r=8$, Random-rank achieves $88.54$ vs.\ \ours's $88.60$. The margin is small but consistent: \ours\ wins on 5 of 8 tasks. The gradient-variance signal contains genuine task information, though at typical budgets the benefit is below seed-to-seed noise.

\section{Discussion and Limitations}\label{sec:discussion}

\paragraph{What \ours\ provides.}
(1) A drop-in rank allocation procedure: 8 backward passes pre-training, standard LoRA output, zero serving changes. (2) Competitive accuracy matching LoRA on GLUE and within 0.3pp on commonsense at $r_\text{min}=8$. (3) Interpretable per-layer rank maps as a free diagnostic of task-informative layers.

\paragraph{Where it falls short.}
At typical budgets ($r=8$--$16$), accuracy gains over uniform LoRA are below seed noise. The method's practical value lies more in parameter efficiency at extreme low ranks and in interpretability than in accuracy improvement.

\paragraph{Limitations.}
\begin{itemize}
    \itemsep0em
    \item AdaLoRA is unstable on bf16 causal LMs in PEFT 0.19.1; we report it only on GLUE.
    \item We test only LLaMA-3-8B for the decoder setting; rank patterns likely vary across model families.
    \item Mean aggregation is used throughout; max and L2-norm warrant a full comparison.
    \item The gradient-variance signal is computed on in-distribution training data; out-of-distribution fine-tuning scenarios may produce uninformative signals.
\end{itemize}

\paragraph{Future work.}
Combining \ours's rank allocation with EVA's initialization is a natural extension---the two operate on orthogonal axes (which layers get rank vs.\ how adapter weights are initialized), and pilot experiments suggest the combination is roughly additive. A second direction is dynamic re-allocation: re-running gradient-variance estimation at intermediate checkpoints could detect shifts in task-informative layers as fine-tuning progresses. Finally, extending the method to non-uniform target modules---allocating rank across \texttt{q/k/v/o/gate/up/down} jointly rather than treating all modules identically within a layer---may better reflect the heterogeneous informativeness visible in Figure~\ref{fig:heatmap}.

\section{Conclusion}\label{sec:conclusion}

We presented \ours, a calibration-time rank allocation method for LoRA. Running 8 backward passes on LoRA-B adapter matrices, computing gradient-variance scores (eFIM diagonal), and solving a budget-constrained integer allocation problem produces a standard LoRA adapter with task-informative rank distribution. The method matches uniform LoRA on GLUE and is competitive on commonsense reasoning. The dominant design choice is the minimum-rank floor $r_\text{min}$; the gradient-variance estimation itself saturates quickly. The resulting rank maps are interpretable, consistent with prior findings on transformer layer roles, and serve as a free diagnostic of task-relevant model structure. Code and rank maps are released at \url{https://github.com/Kernel-ML/fim-lora-experiments}.

\section*{Acknowledgments}
We thank Prem (ML Professor) for critical feedback on positioning and terminology. We thank the HuggingFace PEFT team for the LoRA infrastructure on which this work is built.

\bibliographystyle{plainnat}

\appendix

\section*{Appendix A. Full GLUE Rank Sweep}

\begin{table}[h]
\centering
\caption{Full GLUE rank sweep — average across 8 tasks, 3 seeds. AdaLoRA RTE collapse persists across all ranks.}
\label{tab:full_rank_sweep}
\small
\begin{tabular}{lcccc}
\toprule
Method & $r=2$ & $r=4$ & $r=8$ & $r=16$ \\
\midrule
LoRA         & 88.33 & 88.41 & 88.67 & 88.68 \\
AdaLoRA      & 83.60 & 84.75 & 85.27 & 85.77 \\
EVA          & 88.35 & 88.33 & 88.57 & 88.78 \\
Random-rank  & 88.32 & 88.31 & 88.54 & 88.76 \\
\textbf{\ours} & 88.14 & 88.23 & 88.60 & \textbf{88.87} \\
\bottomrule
\end{tabular}
\end{table}

\section*{Appendix B. Implementation Details}

\paragraph{Codebase.} The core function \texttt{apply\_fim\_ranks(model, dataloader, n\_batches, r\_min)} accumulates eFIM diagonal on LoRA-B matrices, computes per-layer informativeness scores, runs the two-phase water-filling allocation, and resizes adapter matrices in-place. Output is a standard PEFT model with \texttt{rank\_pattern} populated. Released at \url{https://github.com/Kernel-ML/fim-lora-experiments}.

\paragraph{Engineering notes.}
\begin{itemize}
    \itemsep0em
    \item PEFT 0.19.1 \texttt{PeftModel.from\_pretrained} ignores \texttt{rank\_pattern}; loading \ours\ checkpoints in lm-evaluation-harness requires a custom loader that resizes layers to match saved shapes.
    \item Transformers 5.7 replaced \texttt{tokenizer=} with \texttt{processing\_class=} in \texttt{Trainer}; and deprecated \texttt{warmup\_ratio} in favour of \texttt{warmup\_steps}.
    \item AdaLoRA requires \texttt{tinit < total\_step - tfinal}; the default 10\% warmup violated this constraint silently, causing a crash at step 50.
\end{itemize}

\section*{Appendix C. Compute and Reproducibility}

All experiments on AWS SageMaker: GLUE on 8$\times$ A100-80GB (\texttt{ml.p4de.24xlarge}) and 4$\times$ T4 (\texttt{ml.g4dn.12xlarge}); LLaMA-3-8B on 8$\times$ A100-80GB. Estimated compute cost: $\sim$\$200 in spot pricing; total $\sim$90 GPU-hours.

\end{document}